# Network Wide Evacuation Traffic Prediction in a Rapidly Intensifying Hurricane from Traffic Detectors and Facebook Movement Data: A Deep Learning Approach

Md Mobasshir Rashid,[1] Rezaur Rahman, Ph.D.,[2] and Samiul Hasan, Ph.D. [3]

[1]Graduate Research Assistant, Department of Civil, Environmental, and Construction Engineering, University of Central Florida, Orlando, FL, 32816; email: mdmobasshir.rashid@ucf.edu (Corresponding author)

[2]Senior Data Scientist, Amadeus, Dallas, TX, 75252; email: rezaur.rahman@knights.ucf.edu

[3]Associate Professor, Department of Civil, Environmental, and Construction Engineering, University of Central Florida, Orlando, FL, 32816; email: samiul.hasan@ucf.edu

**ABSTRACT:**

Traffic prediction during hurricane evacuation is essential for optimizing the use of transportation infrastructures. It can reduce evacuation time by providing information on future congestion in advance. However, evacuation traffic prediction can be challenging as evacuation traffic patterns is significantly different than regular period traffic. A data-driven traffic prediction model is developed in this study by utilizing traffic detector and Facebook movement data during Hurricane Ian, a rapidly intensifying hurricane. We select 766 traffic detectors from Florida's 4 major interstates to collect traffic features. Additionally, we use Facebook movement data collected during Hurricane Ian's evacuation period. The deep-learning model is first trained on regular period (May-August 2022) data to understand regular traffic patterns and then Hurricane Ian's evacuation period data is used as test data. The model achieves 95% accuracy (RMSE = 356) during regular period, but it underperforms with 55% accuracy (RMSE =

1084) during the evacuation period. Then, a transfer learning approach is adopted where a pretrained model is used with additional evacuation related features to predict evacuation period traffic. After transfer learning, the model achieves 89% accuracy (RMSE = 514). Adding Facebook movement data further reduces model's RMSE value to 393 and increases accuracy to 93%. The proposed model is capable to forecast traffic up to 6-hours in advance. Evacuation traffic management officials can use the developed traffic prediction model to anticipate future traffic congestion in advance and take proactive measures to reduce delays during evacuation.

**PRACTICAL APPLICATIONS:**

Hurricane evacuation causes significant traffic congestion in transportation networks. Increased traffic demand can affect evacuation process as it delays the movement of people to safer locations. To remedy this issue, an accurate traffic prediction model is beneficial for evacuation traffic management. The prediction model can give expected traffic volume on evacuation routes well in advance which will allow traffic management agencies to prepare for and activate strategies such as emergency shoulder utilization, adjustments to signal timing for optimal traffic flow etc. on those evacuation routes. This work aims to construct a data-driven model for the purpose of predicting traffic flow with a lead time of up to 6 hours. The model can be used to make network-wide traffic forecasting in real time. Thus, practitioners can use this tool to effectively implement evacuation traffic management strategies by determining the timing, locations, and extent of those strategies based on predicted traffic volume. Another benefit of this model is that it can be trained with data from normal period and historical hurricane evacuations and then be implemented for future hurricanes.

**INTRODUCTION:**

In recent years, coastal residents of the United States have experienced significant adverse effects from the occurrence of major hurricanes, including but not limited to Hurricanes Irma, Ida, Ian, Harvey, and Idalia. Hurricane events are getting alarmingly intense and frequent along US east coast, mostly due to global climate change (Knutson et al., 2022). Consequently, residents of the United States living near the coast are more likely to be hit by hurricanes, especially during rapidly intensifying hurricanes when people have less time to respond. Major hurricanes can be devastating and cause severe property damage and loss of lives (*Cost of Natural Disasters*, 2017). To mitigate such effects and save lives, emergency managers employ evacuation orders based on the time and location of the hurricane landfall. Although hurricane evacuation plays a vital role to save vulnerable population, evacuation traffic creates sudden demand surge causing traffic congestion and other issues such as increase in crashes and delays in reaching shelters. For example, about 6.5 million Floridians were ordered to leave their homes during Hurricane Irma (Viswanathan, 2021). This caused major congestions and major accidents on I-75 and I-95 (Rahman, Bhowmik, et al., 2021; Rahman, Hasan, et al., 2021). Several traffic management techniques are used to reduce heavy traffic, such as using the shoulder, allocating contraflow, giving clear instructions for evacuation routes, and so on. (Murray-Tuite & Wolshon, 2013). However, to efficiently manage evacuation, traffic managers need to understand the prevailing and future traffic condition of the network. A network-level traffic prediction model can assist emergency management for a proactive application of such strategies to efficiently manage evacuation traffic.

However, predicting evacuation traffic is more difficult due to uncertain demand variations from regular period traffic. Evacuation traffic pattern does not show any peaking (morning and evening peak) behavior like regular traffic. Moreover, during evacuation period the road network operates at its capacity for a prolonged period resulting in stop and go traffic conditions. Additionally, a sudden change of hurricane path can cause changes in evacuation orders. As a result, it induces sudden demand surge on different evacuation routes. Traditionally, mathematical modeling and simulation methods have been used to forecast evacuation traffic (Barrett et al., 2000; Chen et al., 2020; Q. Li et al., 2006). These approaches rely on mathematical assumption-based user-equilibrium solutions to estimate network wide traffic, which might not hold true during an evacuation period. Moreover, lack of input from real-time traffic data makes these approaches less robust against sudden demand surge. To address this issue, data-driven methodologies can be used. Data-driven models can forecast future traffic by analyzing historical evacuation traffic patterns and traffic data. These models do not rely on any assumptions about the behavior of evacuees.

Traditionally data-driven traffic prediction models are formulated as simple time series problems. In such approaches, various models such as ARIMA, SARIMA, SARIMAX, K Nearest Neighbor (KNN), Support Vector Regression (SVR), Decision Tree, Artificial Neural Network (ANN) models have been widely used (Ahn et al., 2016; Brian Smith & Demetsky, 1997; Cai et al., 2016). However, spatiotemporal time-series prediction problems are more complicated since they require capturing both spatial and temporal correlation among traffic variables (Jiang & Luo, 2022). Such traffic prediction problems require models which can deal with higher dimensionality of traffic variables. Hence, traditional modeling approaches are not suitable for spatiotemporal traffic prediction. Recent developments in machine learning have pushed the boundary of traditional ANN models and enabled learning of high dimensional data through multilayered parameters known as deep learning. Deep learning techniques such as Convolutional Neural Networks (CNNs) and Graph Convolutional Neural Networks (GCNNs) can learn spatiotemporal variations in traffic patterns to forecast future traffic; providing more accurate predictions compared to traditional models (Ahn et al., 2016; Cai et al., 2016; Innamaa, 2005).

This study employs the Graph Convolutional Neural Network (GCNN) architecture to forecast network-wide traffic volume during evacuations. GCNN models acquire knowledge of the transportation network by conceptualizing it as a graph, in which road intersections are represented as nodes and roads that connect those intersections are regarded as edges. These models are structured to learn traffic state such as intersection level traffic volume, link travel time or speed of an entire network via a graph convolution layer. The graph convolution layer utilizes a graph theoretic approach to extract spatial cross correlations among input features (Wu et al., 2021). Many studies have applied different graph representation techniques to achieve cutting edge traffic prediction accuracy (Wu et al., 2019). However, one of the limitations to train such large-scale models is that they need extensive data. Hence, most of the previous studies applied GCNN models for regular traffic conditions where they have enough data to train the model. On the contrary, in our approach we create the model to forecast traffic throughout the evacuation period of a rapidly intensifying hurricane. Since the evacuation period in such events lasts for only 1 to 3 days, available data is not sufficient to train such large-scale models. To overcome this issue, in this study we adopt a transfer learning approach similar to what was proposed in (Rahman & Hasan, 2023).

We utilize a Dynamic Graph Convolution Neural Network (DGCNN) based deep learning architecture to forecast network-wide evacuation traffic for Hurricane Ian's evacuation period. We call our model 'dynamic' because it utilizes variations of travel time at each time step to understand the congestion propagation at the whole network. On September 28, 2022, Hurricane Ian made landfall in Florida. It was a rapidly intensifying hurricane, and the evacuation orders were placed just two days (September 26-27, 2022) before the hurricane. So, we have data of only two days from the hurricane evacuation period to train the model. Due to data scarcity of evacuation period, we train our model with regular period data (May 15– August 15, 2022) and test the model performance for evacuation period traffic.

For training the model, we use data from two sources. Like previous studies, we use traffic data from roadway detectors. The Florida Department of Transportation (FDOT) maintains Microwave radar

Vehicle Detection System (MVDS) detectors on the interstate roads in Florida. These detectors provide traffic data such as speed, volume, occupancy at high spatiotemporal resolutions (Ghorbanzadeh et al., 2021; Rahman, Roy, et al., 2021). Moreover, to improve the accuracy of the model and capture the variations in evacuation traffic demand, we also use social media data as an input to the model. We use mobility data from 'Data for Good at Meta' platform which provides movement between places during crisis events such as hurricanes (*Data for Good at Meta*, 2022). The data consists of aggregated real-time movements of Facebook users between different administrative levels at 8-hour intervals. We develop a data processing tool to integrate this mobility data with traffic detector data for the entire interstate network of Florida.

We develop the prediction model focusing on the hypotheses that there is a strong correlation between the number of evacuees (from social media data) and change in surge of traffic demand (from traffic detector data) in major freeways during evacuation period. The main contributions of this study are as follows:

i. It develops a deep learning-based traffic prediction model that can accurately represent the spatiotemporal dynamics of evacuation traffic for a rapidly intensifying hurricane;

ii. It identifies the challenges and develops methods to process discontinuous Facebook movement data to reveal real-time evacuation travel demand variations; and

iii. It demonstrates the utility of Facebook movement data to improve the accuracy of spatiotemporal traffic prediction model; to the best of our knowledge Facebook movement data was never considered in traffic prediction models.

**LITERATURE REVIEW:**

Evacuation studies used statistical patterns to analyze individual evacuation behaviors (Dow & Cutter, 1998; DRABEK, 1992). Later, different discrete choice models were used to determine contributing factors to people's evacuation decisions (Lindell et al., 2018; Murray-Tuite & Wolshon, 2013; Wong et al., 2018). Researchers previously focused on analyzing factors that lead to evacuation decisions (Fry & Binner, 2016; Hasan et al., 2011, 2013), mobilization time (Sadri et al., 2013), departure time (Pel et al., 2012), destination choice (Mesa-Arango et al., 2013; Wilmot, 2006), evacuation mode and destination type (Bian et al., 2019), evacuation plan adaptation (Bian et al., 2022) etc. Insights from evacuation behavior studies can also benefit evacuation traffic modeling studies. For example, behavioral studies can provide information about potential evacuation routes, departure time or evacuation destinations. Previously, several studies used mathematical or simulation-based frameworks to model evacuation traffic demand. For example, Chen et al. (2020) developed a simulation-based framework to predict evacuation traffic due to wildfire. Other studies used different optimization techniques to increase evacuation efficiency (Shahabi & Wilson, 2018).

There are several limitations of evacuation behavior studies. They mainly utilized survey data which are expensive and may not represent the overall population. Also, survey-based studies do not perform well in traffic prediction models due to low sample size. Additionally, simulation-based studies use several assumptions on population behavior which may not capture well actual evacuation traffic demand.

Social media data can be used to overcome limitations of traditional evacuation traffic prediction approaches. They provide geotagged posts which can provide population density during evacuations. We can also get information about traffic congestions which can be integrated to transportation network for better traffic prediction. Previous studies used social media data to detect natural disasters (Kryvasheyeu et al., 2015) and modeling human mobility (Roy et al., 2019). Another study extracted evacuation behavior from Twitter for traffic prediction at the level of a road segment (Roy et al., 2021). Although previous studies used Twitter to analyze evacuation behavior, Twitter data has several limitations such as it lacks representativeness and cannot provide any meaningful variable that can be readily used for any traffic prediction models (Carley et al., 2016). Additionally, researchers used various filtering algorithms to extract information from raw Twitter data, potentially introducing biases in model results (Wang & Ye, 2018).

Alternatively, population data provided by Facebook's 'Data for Good at Meta' platform can be used to analyze population distribution during evacuation. This data provides several useful metrics such as 'z-score' that can be used to identify hot spots of population distribution during crisis events (Jia et al., 2020). Facebook also provides movement data which illustrates anonymous people movement between administrative regions during crisis events. The movement data collects movements for those users who turn on their device's GPS location at 8-hour intervals. This data provides highly granular information of evacuation dynamics which Twitter data cannot provide. Also, movement data provides important insights which can be beneficial for modeling evacuation demand. This data has not been previously used to predict evacuation traffic.

Recent developments of high computational power have enabled researchers to use different deep learning models such as Long Short-Term Memory (LSTM) model, Convolutional Neural Network (CNN), Graph Convolutional Neural Network (GCNN), or a hybrid approach of integrating those models such as CNN-LSTM or GCNN-LSTM models etc. to predict traffic state with higher accuracy (Cui et al., 2020; Y. Li et al., 2018; Zhao et al., 2020). Jiang & Luo (2022) concluded that Graph Neural Network (GNN) based models are becoming popular in traffic prediction studies. Majority of these GNN models were developed to predict traffic in regular conditions. They cannot be used for evacuation traffic prediction because there is a significant difference between regular traffic and evacuation traffic (Rahman, Hasan, et al., 2021). Recently, Rahman & Hasan (2023) proposed a DGCN-LSTM model to predict evacuation traffic during Hurricane Irma considering traffic detector data. However, they didn't use any social media data as an input feature that would represent evacuation demand.

In summary, existing deep learning-based traffic prediction models considered only traffic detector data to predict regular period traffic states. Few studies considered network-level evacuation traffic dynamics by using traffic detector data. In this study, we propose a methodology that combines both detector data and Facebook movement data to predict evacuation traffic for a rapidly intensifying hurricane. The methodology can be applied to all emergency events where evacuation period lasts for a short period of time.

**DATA DESCRIPTION**

**Traffic Detector Data**

In this study, we collected hourly traffic detector data from Regional Integrated Transportation Information System (RITIS) which provides real-time information on traffic speed, volume, occupancy at a high resolution (*RITIS*, 2022). We selected four major interstates of Florida: I-95 (northbound), I-75 (northbound), I-4 (eastbound) and Florida's turnpike (northbound) based on analyzing the major evacuation routes from previous hurricanes in Florida (Rahman, Roy, et al., 2021). We collected regular period traffic detector data from May 15 – August 15, 2022, and evacuation period traffic detector data of Hurricane Ian from September 26 – September 27, 2022. We processed the raw data to discard detectors with missing data, zero values etc. The details of data processing are discussed in the Methodology section. After data processing was done, we selected 766 detectors to construct the graph network. **Fig. 1** shows the location of detectors after data processing.

**Facebook Movement Data**

We also extracted Facebook movement data from "Data for Good at Meta" platform that provides the number of people moving between administrative regions at 8-hour intervals during Hurricane Ian's evacuation period. The data includes Facebook users' mobility information for **two separate 8-hour**

**periods** 3 am-11 am and 11 am-7 pm during Hurricane Ian's evacuation period. Facebook did not provide any movement data for 7 pm-3 am in each day. The dataset also included baseline movement data between tiles, the baseline period data was collected 45 days before the movement map was first generated (*Data for Good at Meta*, 2022). The extracted data include users' movements between small geospatial tiles of Bing tile level 14, where each tile is about 2.4 × 2.4 km at the equator (Maas et al., 2019). We assumed that most people who evacuated using freeways must travel longer distance than what was provided in tile-level movement data. So, we aggregated tile-level movements to county subdivision level movements. Details about the movement data processing are described in the Methodology section.

Hurricane Ian made landfall on the west coast of Florida with 12 counties on the western coast issued mandatory evacuation orders on September 27, 2022 (Ian Evacuation, 2022). **Fig. 2** shows evacuation zones under mandatory evacuation orders along with respective zone level during Hurricane Ian. Zone level A denotes a high-risk zone while zone level E indicates a low-risk zone. Majority of these evacuation zones were in the west coast as Hurricane Ian was predicted to hit Florida from the west coast.

We compared percent increase of movement patterns in the evacuation period compared to the baseline period. The baseline period is 45 days before the movement data is first generated. Charlotte, Pinellas, Pasco, Hillsborough, and Sarasota counties issued evacuation orders on September 26. All of these counties are situated in the west and southwest coast of Florida. We observed a significant increase of users' movements on September 26. Majority of these movements generated from Central Florida, west and southwest coasts of Florida. People moved from west and southwest coast which includes major cities such as Tampa to Central Florida region. **Fig. 3** shows the movement patterns starting at 3 am and ending at 11 am of September 26 (2 days before landfall time). From the

Facebook movement data, we observed that movements increased in Central Florida, southeast and west coast regions compared to baseline movements.

**Fig. 4** shows movement patterns during September 27 (3 am-11 am). Evacuation orders were already placed for 12 counties during this time period. We observed high movements in Central Florida region. But percent change of movements decreased in the west coast compared to previous day's movements. It indicates that population in west coast also decreased on September 27 as people started evacuating from the region.

Based on our analysis we found that majority of people evacuated from the western region to the Central Florida region. In regular traffic condition, Interstate I-4 Eastbound serves majority of traffic traveling from west coast to Central Florida region. To correlate the movement data with traffic volume data, we plot cumulative traffic volume for an eastbound detector of Interstate I-4; the detector is placed close to Central Florida (**Fig. 5**). We used cumulative hourly traffic flow instead of hourly traffic flow to compare whether I-4 Eastbound dealt with higher amount of traffic flow during the evacuation period. We chose 8-hour timeframe to match with the 8-hour timeframe of Facebook movement data as showed in **Fig. 3** and **Fig. 4**. We calculated mean traffic flow based on hour and day of the week from regular period traffic data (May 15 – August 15, 2022). Then we extracted cumulative baseline period flow of the selected detector. We found that during September 26, 2022, cumulative flow was around 25,000 vehicles and it reached around 50,000 vehicles on September 27, 2022. Facebook movement data shows increased movements in Central Florida during evacuation period, and traffic detector data also illustrates higher crisis period flow over baseline period flow. Additionally, many people evacuated from west coast to Central Florida through Interstate I-4 Eastbound, which explains increase in cumulative flow compared to baseline period cumulative flow. The human movement patterns from Facebook followed similar trends of the actual traffic movement during Hurricane Ian.

## METHODOLOGY

**Detector Data Processing**

Raw traffic detector data is prone to errors due to detector malfunctioning, bad weather, duplicate or missing entries, wrong storing, etc. During hurricane evacuation period, majority of vehicle face 'stop and go' traffic congestion in major freeways, which detectors may fail to capture (Rahman & Hasan, 2023). To address these issues, we performed extensive data cleaning to prepare final training data for the deep-learning model. First, we removed detectors having missing values higher than 20% of the observations and zero values higher than 40% of the observations. Second, we discarded those detectors having traffic flow per hour per lane (vphpl) higher than 2500 (Rahman & Hasan, 2023). Finally, we applied multivariate iterative imputation to fill up missing values (Pedregosa et al., 2011). **Fig. 6** provides the data-processing steps of traffic detector data.

**Facebook Movement Data Processing**

We followed several steps to process the Facebook users' movement data. Although the movement data were provided at a small geospatial tile level, we assumed that people evacuated further than a tile distance. So, we aggregated the movement volume in each tile's corresponding county subdivision level. There were two types of movement: evacuation within a county subdivision (origin and destination of the movement falls within the same county subdivision) and evacuation between county subdivisions (origin and destination are different county subdivisions). We discarded movements occurring within the same subdivision considering that these movements might not use highways. Then, we assigned closest traffic detectors from RITIS for each movement's origin and destination county subdivisions. We assigned the closest detector based on the minimum distance between the centroid of a county subdivision and selected 766 detectors from RITIS. Then we checked whether the origin and destination traffic detectors of a movement match; if so, we assumed that this movement was less likely to use any highways to evacuate; this was mainly reflecting the movement inside a county subdivision. For each detector, we aggregated the number of inflow and outflow values of Facebook users. The processed data contains aggregate inflow and outflow for 8-hour intervals. We disaggregated them to 1-hour movement by applying an hourly factor calculated from the traffic flow data. **Eq. 1** shows the formula for estimating the hourly factor.

$$hourly\ factor = \frac{F_z}{\sum_{z=1}^{8} F_z} \quad (1)$$

Where $F_z$ indicates total traffic flow of 766 detectors in 1-hour. $z$ denotes the hour ranging from start to end of the 8-hour interval. Then, we multiplied the 8-hour Facebook movement data with the hourly factor to disaggregate it to hourly movement data as shown in **(Eq. 2)**. **Fig. 7** shows the workflow used to process the Facebook movement data.

$$hourly\ movement\ data = Facebook\ movement\ data \times hourly\ factor \quad (2)$$

**Problem Formulation**

In this study, we predict network-level evacuation traffic flow given that traffic volume and evacuation demand data are available at a higher spatiotemporal resolution. To solve this problem, we adopted similar approach of Dynamic Graph Convolution Neural Network and Long Short Term Memory **(DGCN-LSTM)** model developed in (Rahman & Hasan, 2023). We used two stacked layers in our deep learning model. In the first layer, we used dynamic graph convolution (DGCN) operation to capture the spatial cross correlation among traffic state related features. In the dynamic graph convolution approach, weights of the graph were assigned based on changes in travel time between two detectors at each timestep. In the second layer, we used a LSTM unit to capture the temporal dependency among the input features.

Let, $X_t$ be the input features coming from both traffic detector and Facebook movement data. We considered the transportation network as a graph and each traffic detector as a node. The graph is represented as $\mathcal{G}_t(v, e, A_t)$ where $v$ is the set of all nodes (detectors), $e$ denotes the set of edges between detectors (road segment between two detectors) and $A_t$ denotes the weighted adjacency matrix. In our prediction problem, we learn a function $\mathcal{F}$ that takes $l$ instances of input sequences $([X_{t-l}, X_{t-l+1}, \ldots, X_t])$ and predicts future traffic flow $(F_{t+1}, \ldots, F_{t+p})$ for $p$ instances. We can define the problem via **Eq. 3.**

$$\mathcal{F}([X_{t-l}, X_{t-l+1}, \ldots, X_t]; [\mathcal{G}_{t-l}(v, e, A_{t-l})]) = [F_{t+1}, \ldots, F_{t+p}] \quad (3)$$

We used weighted adjacency matrix instead of regular adjacency matrix to make the graph dynamic. The weighted adjacency matrix contains information of change in travel time between detector pairs for each time step. The travel time depends on the speeds of two consecutive detectors at each timestep. Since the weighted adjacency matrix is dynamic and function of travel time, the prediction model can learn the network-level congestion propagation with respect to changes in travel time. The equation of weighted adjacency matrix $A_t$ is defined by **Eq. 4.**

$$A_t(i,j) = \begin{cases} tt_t(i,j), & if\ i \neq j \\ 0 & if\ i = j \end{cases} \quad (4)$$

Equation of travel time, $tt_t(i,j)$ depends on distance $(d^{i,j})$ and speed of two consecutive detectors $(s_t^i, s_t^j)$ at each time step as shown in **Eq. 5**.

$$tt_t(i,j) = \frac{d^{i,j}}{\frac{s_t^i + s_t^j}{2}} \qquad (5)$$

The proposed framework of the **DGCN-LSTM** model is illustrated in **Fig. 8.** The model takes $l$ = 6 hours of sequential data as input and predicts traffic volume for next $p$ = 6 hours.

**Transfer Learning Approach**

To train the proposed DGCN-LSTM model, a substantial amount of input data is required. During regular period traffic prediction task, RITIS can provide high amount of traffic flow data. As a result, the proposed model can predict regular period traffic efficiently. But the goal of this study is to predict traffic flow during evacuation period. As the evacuation process lasts for a short period of time (2 to 5 days) during a rapidly intensifying hurricane, the DGCN-LSTM model cannot be trained with sufficient data. To overcome this issue, we adopted a transfer learning approach (Zhuang et al., 2021).

We first trained the DGCN-LSTM model with regular period data. Then we applied transfer learning approach to predict for evacuation period. However, there is a significant difference between regular and evacuation period traffic. The traffic demand can increase significantly due to evacuation process. Additionally, evacuation period doesn't show any regularity in traffic patterns. To overcome this issue, we extracted only necessary information such as transportation network connectivity and how traffic flow propagates along the network through all detectors by using transfer learning.

The transfer learning approach is divided into 4 parts. The first part is the pretrained DGCN-LSTM model with regular data. We used this model to predict evacuation period traffic. Then, the second part consists of a LSTM model where we trained the LSTM model with evacuation traffic state features along with evacuation demand related features such as distance between a detector and nearest evacuation zone, time left before hurricane landfall, and cumulative population placed under mandatory evacuation orders. The third part is called control layer which is a neural network block with sigmoid activation function. The control layer controls necessary information coming from the output of the DGCN-LSTM model via sigmoid activation function. The fourth layer is called output layer which adds output of second and third layer together to provide final output of evacuation period traffic. Details about the transfer learning approach are described in (Rahman & Hasan, 2023). **Fig. 9** illustrates the transfer learning approach to predict evacuation period traffic. Orange boxes indicate four parts of the transfer learning model, blue boxes indicate input features, and the green box indicates final output of evacuation traffic flow.

**Methodology to Handle Discontinuous Facebook Movement Data**

Facebook provided daily movement data for 16 hours instead of 24 hours due to technical issues. Although we had traffic detector data for 24 hours, we didn't use detector data from 7 pm to 3 am of next day. We selected RITIS detector dataset for following time periods: September 26 (3 am – 7 pm) and September 27 (3 am – 7 pm) to match them with available Facebook movement data. As a result, our sample size for each day reduced to 16 hours. To handle such discontinuity in the dataset, we adopted an indexing approach where we gave an index number for each sequential 6-hour observations in our dataset. As the proposed model takes 6 hours of input data, we maintained the temporal sequence for 6 hours under each index number in the final dataset. For example, an index contains traffic state observations for 5 am, 6 am, 7 am, 8 am, 9 am, and 10 am as input features of the DGCN-LSTM model. The next index contains observations from 6 am – 11 am. Then, we run the prediction model for 10 times. In each iteration, the index numbers of the training, validation and testing dataset observations are randomly shuffled. As a result, we have different index values containing 6-hour observations for training, validation, and testing purposes. The model wasn't trained sequentially on data from the start to the end of the total observation period. As a result, the missing 8-hour timeframe of each day did not affect the model's learning process of how traffic flow propagates in the whole network. By adopting this randomly shuffled indexing technique, the model becomes more robust against data discontinuity.

**INPUT FEATURES**

We used features shown in **Table 1** as input to the prediction model. We used several non-evacuation related features to train the prediction model for regular period. We divided 16-hour time period into 4 different periods: **Early Morning** (3 am-7 am), **Morning** (7 am-11 am), **Mid-day** (11 am-3 pm), **Evening** (3 pm-7 pm). Additionally, we used previous day's mean and standard deviation of traffic flow, previous period's mean and standard deviation of traffic flow, weekday/weekend, and mean traffic speed. For evacuation period traffic prediction, we used evacuation demand related features such as the time left before landfall, distance from the nearest evacuation zones for each detector, and cumulative population under mandatory evacuation orders in the transfer-learned DGCN-LSTM model. We collected declaration time of evacuation orders from different County Emergency Managements' official Twitter posts and counted total number of people living in respective County's evacuation zones to generate 'population under mandatory evacuation order' variable. We also used human movement inflow and outflow values for each detector from the Facebook movement data.

**RESULTS**

**Regular Period Traffic Prediction**

We implemented the prediction model by using Python's Pytorch environment (*Pytorch*, 2016). For regular period traffic prediction, we discarded evacuation demand related features. We used 90% data for training, 5% for validation, and 5% data to test model performance. To train the model, we used ADAM optimizer and assigned mean squared error as loss functions. To compare different models' performance, we considered several loss criteria such as Root Mean Square Error **(RMSE)**, Mean Absolute Error **(MAE)**, Mean Absolute Percentage Error **(MAPE)**, and $R^2$ value. Equations for loss criteria are provided in **(Eq. 6 – 8)**. Here, $F_{actual,i}$ denotes the actual traffic flow at timestep $i$, and $F_{predicted,i}$ is the predicted traffic flow at timestep $i$.

$$RMSE = \sqrt{\frac{1}{N}\sum_{i=1}^{N}(F_{actual,i} - F_{predicted,i})^2} \qquad (6)$$

$$MAE = \frac{1}{N}\sum_{i=1}^{N}|F_{actual,i} - F_{predicted,i}| \qquad (7)$$

$$MAPE = \frac{1}{N}\sum_{i=1}^{N}\left|\frac{F_{actual,i} - F_{predicted,i}}{F_{actual,i}}\right| \qquad (8)$$

For regular period traffic prediction, we compared the performance of proposed DGCN-LSTM model against several baseline models such as LSTM, CNN-LSTM, and GCN-LSTM. We first trained all models without adding the Facebook movement data. All models achieved similar $R^2$ values (95%). DGCN-LSTM model achieved lowest RMSE values of 356.47 and MAPE of 8.74 compared to other baseline models. Then, we extracted baseline movement by considering day of the week and hour to generate inflow and outflow values for all detectors in the regular period. After adding Facebook baseline movement data, the RMSE values further decreased for all models. DGCN-LSTM model outperformed other baseline models with the lowest RMSE values of 319.91 and MAPE of 7.46. **Table 2** presents average model performances after running 10 times and impact of Facebook movement data on prediction performance. DGCN-LSTM outperformed other baseline models during regular period, and addition of Facebook movement data increased all models' performances.

**Evacuation Period Traffic Prediction**

In case for evacuation period traffic prediction, we first used only non-evacuation related features to predict evacuation traffic. Like regular period traffic prediction, we compared the effect of adding Facebook movement data on all models' performances. We used 80% data for training, 10% for validation and 10% data to test model performance. We used ADAM optimizer and mean squared error as loss function. **Table 3** shows average model performances after running 10 times during evacuation period. As shown in **Table 3**, all models performed poorly. However, the result is intuitive since regular period traffic patterns differ significantly from evacuation period traffic. RMSE values for all models were more than 1000 even after adding the Facebook movement data. However, the DGCN-LSTM model still performed better than other models in this scenario with RMSE value of 1084.36 and $R^2$ value of 0.55. Moreover, after adding the Facebook data, overall performance of all models slightly improved. The RMSE value of the LSTM model decreased to 1325.69 from 1328.72; the $R^2$ value of LSTM also increased from 32% to 33%. The RMSE value of CNN-LSTM model decreased from 1180.63 to 1134.51, and $R^2$ value also increased from 46% to 50%. Similar to previous cases, DGCN-LSTM model trained with Facebook movement data outperformed other models with RMSE value of 1053.24 and $R^2$ value of 0.57.

Next, to improve the DGCN-LSTM model's predictability during evacuation period, we applied a transfer learning approach proposed by (Rahman & Hasan, 2023). We used a pretrained DGCN-LSTM model by using evacuation period traffic state as input. The pretrained model contained information on how the detectors were connected in the transportation network and the traffic flows between different detectors. Additionally, we trained another neural network with **evacuation demand related features** to capture temporal dependency of evacuation traffic. In this modeling architecture, a control layer is used, to control relevant information (such as network connectively, flow propagation pattern etc.,) transfer from regular period traffic to evacuation period traffic. By adopting the transfer learning technique, the RMSE values of DGCN-LSTM model decreased to 514.20 when Facebook data were not used, and the $R^2$ value increased from 0.55 to 0.89 as given in **Table 3**. After adding Facebook movement data to the transfer-learned DGCN-LSTM model, the RMSE value further reduced to 393.28 and the $R^2$ value increased from 0.57 to 0.93. The transfer-learned DGCN-LSTM model trained with Facebook data outperformed other baseline models. **Table 3** illustrates the benefits of using transfer learning approach along with Facebook movement data to predict evacuation period traffic during a rapidly intensifying hurricane with higher accuracy.

**DISCUSSIONS**

In the study, we present a deep learning model to predict traffic during hurricane evacuation. We integrate both traffic detector data and Facebook movement data to the proposed DGCN-LSTM architecture. Both regular period and evacuation period traffic predictions achieve higher accuracy when Facebook movement data are used. After applying the transfer learning approach, the proposed model predicts evacuation period traffic up to 6 hours in advance with 93% accuracy. **Fig. 10** shows the correlation between actual traffic and predicted traffic of transfer learned DGCN-LSTM model when Facebook movement data is not utilized. The model achieves 89% accuracy and RMSE value of 514.2. When Facebook movement data are used, the transfer-learned DGCN-LSTM model learns the evacuation period traffic patterns very well as actual and predicted traffic almost matched with each other. **Fig. 11** shows the correlation between actual and predicted traffic flow when Facebook movement data are utilized.

We also plot the detector wise variations of actual and predicted traffic flows without Facebook movement data as shown in **Fig. 12**. The overall symmetric mean absolute percentage error **(SMAPE)** for different prediction horizon remains less than 12%. When Facebook movement data are used, the SMAPE values for different prediction decreases from 12% to 7% as shown in **Fig. 13**. It indicates that the model captures spatiotemporal patterns of evacuation period traffic very well when Facebook movement data are used along with traffic detector data. The equation of SMAPE is provided in **(Eq. 9).**

$$SMAPE = \frac{1}{N} \sum_{i=1}^{N} \frac{|F_{actual,i} - F_{predicted,i}|}{(|F_{actual,i}| + |F_{predicted,i}|)/2} \qquad (9)$$

The transfer-learned DGCN-LSTM model can also be used to visualize the network-wide congestion propagation. **Fig. 14** shows the actual traffic flow from RITIS from September 26, 4 am – 9 pm. There was heavy traffic flow in I-75, I-4, and I-95 interstates. During this time, several counties ordered evacuation orders which caused higher traffic flow in I-75 and I-4 interstates. The prediction model captures the network-wide traffic flow very well for different prediction horizons as shown in **Fig. 15.** The visualization capabilities of the DGCN-LSTM model will help evacuation traffic managers to take proactive decisions in real time by providing early information of how future traffic congestion may look like in the transportation network.

**CONCLUSIONS**

In this study, we use a deep learning-based traffic prediction model named DGCN-LSTM to predict traffic during a rapidly intensifying hurricane. The proposed model utilizes traffic detector data and Facebook movement data to predict traffic up to 6 hours in advance. The movement data provides spatio-temporal movement distributions at a high spatial resolution. Evacuation movements increased in Central Florida during Hurricane Ian's evacuation period. Movement data also shows that people evacuated from Florida's west coast to the Central Florida region through Interstate I-4 (Eastbound). The movement data contains information regarding increased human movement through the transportation network, and it is representative of actual traffic flow dynamics in the network. DGCN-LSTM model's performance improves significantly by incorporating the information of increased human movement.

Facebook movement data increases the performance of the transfer-learned DGCN-LSTM model from 0.89 to 0.93.

The study also deals with the challenge of data unavailability and illustrates how to develop a traffic prediction model by incorporating discontinuous data via randomly shuffled indexing approach. Traffic management agencies can use the data-driven prediction model trained with Facebook movement data to predict future traffic congestions in advance, take proactive measures to reduce traffic delays and improve the efficiency of evacuation process. As the model incorporates real-time detector and social media data, agencies can also implement it to identify vulnerable zones with high congestion probabilities earlier when hurricane unfolds in real time.

There are several limitations of this study. Due to data scarcity, we only use 16 hours of daily movement data. So, the approach needs more testing with additional data from multiple hurricanes. This way we will be able to develop a more generalized model and improve the robustness of the model against sudden demand surge. We also use cumulative population under mandatory evacuation orders as input features from emergency management officials' Twitter accounts. The model performance should be evaluated if actual population under mandatory evacuation orders can be obtained to obtain more realistic scenario.

**DATA AVAILABILITY STATEMENT:**

The test dataset used in the study is available in the following repository in accordance with funder data retention policies. The train dataset and code that support the findings of this study are available from the corresponding author upon reasonable request.

(https://www.designsafe-ci.org/data/browser/public/designsafe.storage.published/PRJ-4268)

**ACKNOWLEDGEMENTS:**

The authors are grateful to the US National Science Foundation for grants 1917019 and 2122135 to support the research presented in this paper. However, the authors are solely responsible for the findings presented here.

**REFERENCES:**

Ahn, J., Ko, E., & Kim, E. Y. (2016). Highway Traffic Flow Prediction using Support Vector Regression and Bayesian Classifier. *International Conference on Big Data and Smart Computing (BigComp)*, 239–244. https://doi.org/10.1109/BIGCOMP.2016.7425919

Barrett, B., Ran, B., Pillai, R., Barrett, B., & Ran, B. (2000). Developing a Dynamic Traffic Management Modeling Framework for Hurricane Evacuation. *Transportation Research Record*, 115–121. https://doi.org/https://doi. org/10.3141/1733-15

Bian, R., Murray-Tuite, P., Edara, P., & Triantis, K. (2022). Household Hurricane Evacuation Plan Adaptation in Response to Estimated Travel Delay Provided Prior to Departure. *Natural Hazards Review*, *23*(3), 1–15. https://doi.org/10.1061/(asce)nh.1527-6996.0000557

Bian, R., Wilmot, C. G., Gudishala, R., & Baker, E. J. (2019). Modeling household-level hurricane evacuation mode and destination type joint choice using data from multiple post-storm behavioral surveys. *Transportation Research Part C: Emerging Technologies*, *99*(January), 130–143. https://doi.org/10.1016/j.trc.2019.01.009

Brian Smith, B. L., & Demetsky, M. J. (1997). Traffic Flow Forecasting: Comparison of Modeling Approaches. *Journal of Transportation Engineering*, *123*(4). https://doi.org/https://doi.org/10.1061/(ASCE)0733-947X(1997)123:4(261)

Cai, P., Wang, Y., Lu, G., Chen, P., Ding, C., & Sun, J. (2016). A spatiotemporal correlative k-nearest neighbor model for short-term traffic multistep forecasting. *Transportation Research Part C: Emerging Technologies*, *62*, 21–34. https://doi.org/10.1016/j.trc.2015.11.002

Carley, K. M., Malik, M., Landwehr, P. M., Pfeffer, J., & Kowalchuck, M. (2016). Crowd sourcing disaster management: The complex nature of Twitter usage in Padang Indonesia. *Safety Science*, *90*, 48–61. https://doi.org/10.1016/j.ssci.2016.04.002

Chen, Y., Shafi, S. Y., & Chen, Y. fan. (2020). Simulation pipeline for traffic evacuation in urban areas and emergency traffic management policy improvements through case studies. *Transportation Research Interdisciplinary Perspectives*, *7*. https://doi.org/10.1016/j.trip.2020.100210

*Cost of Natural Disasters*. (2017). https://nypost.com/2017/12/22/the-cost-of-natural-disasters-nearly-doubled-in-2017/.

Cui, Z., Ke, R., Pu, Z., & Wang, Y. (2020). Stacked bidirectional and unidirectional LSTM recurrent neural network for forecasting network-wide traffic state with missing values. *Transportation Research Part C: Emerging Technologies*, *118*. https://doi.org/10.1016/j.trc.2020.102674

*Data for Good at Meta*. (2022). https://dataforgood.facebook.com/dfg/about

Dow, K., & Cutter, S. L. (1998). Crying wolf: Repeat responses to hurricane evacuation orders. *Coastal Management*, *26*(4), 237–252. https://doi.org/10.1080/08920759809362356

DRABEK, T. E. (1992). Variations in Disaster Evacuation Behavior: Public Responses Versus Private Sector Executive Decision-Making Processes. *Disasters*, *16*(2), 104–118. https://doi.org/10.1111/j.1467-7717.1992.tb00384.x

Fry, J., & Binner, J. M. (2016). Elementary modelling and behavioural analysis for emergency evacuations using social media. *European Journal of Operational Research*, *249*(3), 1014–1023. https://doi.org/10.1016/j.ejor.2015.05.049

Ghorbanzadeh, M., Burns, S., Rugminiamma, L. V. N., Ozguven, E. E., & Huang, W. (2021). Spatiotemporal analysis of highway traffic patterns in hurricane irma evacuation. *Transportation Research Record*, *2675*(9), 321–334. https://doi.org/10.1177/03611981211001870

Hasan, S., Mesa-Arango, R., & Ukkusuri, S. (2013). A random-parameter hazard-based model to understand household evacuation timing behavior. *Transportation Research Part C: Emerging Technologies*, *27*, 108–116. https://doi.org/10.1016/j.trc.2011.06.005

Hasan, S., Ukkusuri, S., Gladwin, H., & Murray-Tuite, P. (2011). Behavioral model to understand household-level hurricane evacuation decision making. *Journal of Transportation Engineering*, *137*(5), 341–348. https://doi.org/10.1061/(ASCE)TE.1943-5436.0000223

*Hurricane Ian Evacuation Orders*. (2022). https://abcnews.go.com/US/officials-people-evacuate-hurricane-ian/story?id=90931063#:~:text=This%20is%20how%20the%20evacuations,3%20hurricane%2C%20threatening%20coastal%20communities.

Innamaa, S. (2005). Short-term prediction of travel time using neural networks on an interurban highway. *Transportation*, *32*(6), 649–669. https://doi.org/10.1007/s11116-005-0219-y

Jia, S., Kim, S. H., Nghiem, S. V., Doherty, P., & Kafatos, M. C. (2020). Patterns of population displacement during mega-fires in California detected using Facebook Disaster Maps. *Environmental Research Letters*, *15*(7). https://doi.org/10.1088/1748-9326/ab8847

Jiang, W., & Luo, J. (2022). Graph neural network for traffic forecasting: A survey. *Expert Systems with Applications*, *207*. https://doi.org/10.1016/j.eswa.2022.117921

Knutson, T. R., Sirutis, J. J., Bender, M. A., Tuleya, R. E., & Schenkel, B. A. (2022). Dynamical downscaling projections of late twenty-first-century U.S. landfalling hurricane activity. *Climatic Change*, *171*(3–4). https://doi.org/10.1007/s10584-022-03346-7

Kryvasheyeu, Y., Chen, H., Moro, E., Van Hentenryck, P., & Cebrian, M. (2015). Performance of social network sensors during Hurricane Sandy. *PLoS ONE*, *10*(2). https://doi.org/10.1371/journal.pone.0117288

Li, Q., Yang, X. K., & Wei, H. (2006). Integrating Traffic Simulation Models with Evacuation Planning System in a GIS Environment. *IEEE Conference on Intelligent Transportation Systems, Proceedings*. https://doi.org/doi: 10.1109/itsc.2006.1706805

Li, Y., Yu, R., Shahabi, C., & Liu, Y. (2018). DIFFUSION CONVOLUTIONAL RECURRENT NEURAL NETWORK: DATA-DRIVEN TRAFFIC FORECASTING. *International Conference on Learning Representations (ICLR)*.

Lindell, M. K., Murray-Tuite, P., Wolshon, B., & Baker, E. J. (2018). *Large-Scale Evacuation The Analysis, Modeling, and Management of Emergency Relocation from Hazardous Areas*. CRC Press.

Maas, P., Gros, A., McGorman, L., Alex Dow, P., Iyer, S., Park, W., & Nayak, C. (2019). Facebook disaster maps: Aggregate insights for crisis response & recovery. *Proceedings of the International ISCRAM Conference*, *2019-May*(May 2019), 836–847.

Mesa-Arango, R., Hasan, S., Ukkusuri, S. V., & Murray-Tuite, P. (2013). Household-Level Model for Hurricane Evacuation Destination Type Choice Using Hurricane Ivan Data. *Natural Hazards Review*, *14*(1), 11–20. https://doi.org/10.1061/(asce)nh.1527-6996.0000083

Murray-Tuite, P., & Wolshon, B. (2013). Evacuation transportation modeling: An overview of research, development, and practice. *Transportation Research Part C: Emerging Technologies*, *27*, 25–45. https://doi.org/10.1016/j.trc.2012.11.005

Pedregosa, F., Varoquaux, G., Gramfort, A., Michel, V., Thirion, B., Grisel, O., Blondel, M., Müller, A., Nothman, J., Louppe, G., Prettenhofer, P., Weiss, R., Dubourg, V., Vanderplas, J., Cournapeau, D., Brucher, M., & Perrot, M. (2011). Scikit-learn: Machine Learning in Python. *Journal of Machine Learning Research*, *12*, 2825–2830. http://scikit-learn.org.

Pel, A. J., Bliemer, M. C. J., & Hoogendoorn, S. P. (2012). A review on travel behaviour modelling in dynamic traffic simulation models for evacuations. *Transportation*, *39*(1), 97–123. https://doi.org/10.1007/s11116-011-9320-6

*Pytorch*. (2016). https://pytorch.org/

Rahman, R., Bhowmik, T., Eluru, N., & Hasan, S. (2021). Assessing the crash risks of evacuation: A matched case-control approach applied over data collected during Hurricane Irma. *Accident Analysis and Prevention*, *159*(December 2020), 106260. https://doi.org/10.1016/j.aap.2021.106260

Rahman, R., & Hasan, S. (2023). A deep learning approach for network-wide dynamic traffic prediction during hurricane evacuation. *Transportation Research Part C: Emerging Technologies*, *152*. https://doi.org/10.1016/j.trc.2023.104126

Rahman, R., Hasan, S., & Zaki, M. H. (2021). Towards reducing the number of crashes during hurricane evacuation: Assessing the potential safety impact of adaptive cruise control systems. *Transportation Research Part C: Emerging Technologies*, *128*, 103188. https://doi.org/10.1016/j.trc.2021.103188

Rahman, R., Roy, K. C., & Hasan, S. (2021). Understanding Network Wide Hurricane Evacuation Traffic Pattern from Large-scale Traffic Detector Data. *IEEE Conference on Intelligent Transportation Systems, Proceedings, ITSC*, *2021-Septe*, 1827–1832. https://doi.org/10.1109/ITSC48978.2021.9564480

*RITIS*. (2022). https://ritis.org/intro

Roy, K. C., Cebrian, M., & Hasan, S. (2019). Quantifying human mobility resilience to extreme events using geo-located social media data. *EPJ Data Science*, *8*(1). https://doi.org/10.1140/epjds/s13688-019-0196-6

Roy, K. C., Hasan, S., Culotta, A., & Eluru, N. (2021). Predicting traffic demand during hurricane evacuation using Real-time data from transportation systems and social media. *Transportation Research Part C: Emerging Technologies*, *131*. https://doi.org/10.1016/j.trc.2021.103339

Sadri, A. M., Ukkusuri, S. V., & Murray-Tuite, P. (2013). A random parameter ordered probit model to understand the mobilization time during hurricane evacuation. *Transportation Research Part C: Emerging Technologies*, *32*, 21–30. https://doi.org/10.1016/j.trc.2013.03.009

Shahabi, K., & Wilson, J. P. (2018). Scalable evacuation routing in a dynamic environment. *Computers, Environment and Urban Systems*, *67*, 29–40. https://doi.org/10.1016/j.compenvurbsys.2017.08.011

Viswanathan, K. (2021). *Florida Statewide Regional Evacuation Study Program Regional Behavioral Analysis prepared for Northeast Florida Regional Planning Council*. www.camsys.com

Wang, Z., & Ye, X. (2018). Social media analytics for natural disaster management. *International Journal of Geographical Information Science*, *32*(1), 49–72. https://doi.org/10.1080/13658816.2017.1367003

Wilmot, C. G. (2006). *Modeling Hurricane Evacuation Traffic: Testing the Gravity and Intervening Opportunity Models as Models of Destination Choice in Hurricane Evacuation*. www.ltrc.lsu.edu

Wong, S., Shaheen, S., & Walker, J. (2018). *Understanding Evacuee Behavior: A Case Study of Hurricane Irma*. https://doi.org/10.7922/G2FJ2F00

Wu, Z., Pan, S., Chen, F., Long, G., Zhang, C., & Yu, P. S. (2021). A Comprehensive Survey on Graph Neural Networks. *IEEE Transactions on Neural Networks and Learning Systems*, *32*(1), 4–24. https://doi.org/10.1109/TNNLS.2020.2978386

Wu, Z., Pan, S., Long, G., Jiang, J., & Zhang, C. (2019). Graph WaveNet for Deep Spatial-Temporal Graph Modeling. *Proceedings of the Twenty-Eighth International Joint Conference on Artificial Intelligence (IJCAI-19)*. https://doi.org/https://doi.org/10.48550/arXiv.1906.00121

Zhao, L., Song, Y., Zhang, C., Liu, Y., Wang, P., Lin, T., Deng, M., & Li, H. (2020). T-GCN: A Temporal Graph Convolutional Network for Traffic Prediction. *IEEE Transactions on Intelligent Transportation Systems*, *21*(9), 3848–3858. https://doi.org/10.1109/TITS.2019.2935152

Zhuang, F., Qi, Z., Duan, K., Xi, D., Zhu, Y., Zhu, H., Xiong, H., & He, Q. (2021). A Comprehensive Survey on Transfer Learning. *Proceedings of the IEEE*, *109*(1), 43–76. https://doi.org/10.1109/JPROC.2020.3004555

**FIGURE CAPTION LIST:**

- **Fig. 1.** Selected detector distribution from RITIS
- **Fig. 2.** Mandatory evacuation zones during Hurricane Ian
- **Fig. 3.** (Left) Origins of movements; (right) destinations of movements during September 26 (3am-11am)
- **Fig. 4.** (Left) Origins of movements; (right) destinations of movements during September 27 (3am-11am)
- **Fig. 5.** Eastbound cumulative traffic flow on an I-4 detector
- **Fig. 6.** Processing of traffic detector data
- **Fig. 7.** Processing of Facebook movement data
- **Fig. 8.** Framework of the DGCN-LSTM model
- **Fig. 9.** Transfer learning approach for evacuation period traffic prediction
- **Fig. 10.** Correlation between actual traffic and predicted traffic for 6-hour time horizon (when Facebook movement data is not used)
- **Fig. 11.** Correlation between actual traffic and predicted traffic for 6-hour time horizon (when Facebook movement data is used)
- **Fig. 12.** Detector wise actual flow vs. predicted flow with SMAPE values (without Facebook movement data)
- **Fig. 13.** Detector wise actual flow vs. predicted flow with SMAPE values (with Facebook movement data)
- **Fig. 14.** Congestion propagation visualization of actual traffic flow (vertical color bar denotes traffic flow)
- **Fig. 15.** Congestion propagation visualization of predicted traffic flow (vertical color bar denotes traffic flow)

**TABLES:**

**Table 1.** Input features

| Non-evacuation related features | Evacuation demand related features | Facebook movement features |
|---|---|---|
| Detector Id | Time left before landfall | Human inflow to a traffic detector |
| Time periods (early morning, morning, mid-day, evening) | Cumulative population under mandatory evacuation orders | Human outflow from a traffic detector |
| Weekday or Weekend | Distance from nearest evacuation zones | - |
| Traffic flow at current time t | - | - |
| Previous day mean traffic flow | - | - |
| Previous period mean traffic flow | - | - |
| Previous day standard deviation of traffic flow | - | - |
| Previous period standard deviation of traffic flow | - | - |
| Mean speed over an hour | - | - |

**Table 2.** Model performances for regular period traffic prediction

|  | Without Facebook Data | | | | With Facebook Data | | | |
| --- | --- | --- | --- | --- | --- | --- | --- | --- |
| Model | RMSE | MAE | MAPE | $R^2$ | RMSE | MAE | MAPE | $R^2$ |
| LSTM | 374.76 | 227.58 | 9.53 | 0.95 | 343.29 | 196.24 | 8.45 | 0.96 |
| GCN-LSTM | 374.73 | 220.16 | 9.28 | 0.95 | 343.24 | 195.68 | 8.08 | 0.96 |
| CNN-LSTM | 368.37 | 217.03 | 8.84 | 0.95 | 339.02 | 194.79 | 8.06 | 0.96 |
| DGCN-LSTM | 356.47 | 215.63 | 8.74 | 0.95 | 319.91 | 186.19 | 7.46 | 0.96 |

**Table 3.** Model performances for evacuation period traffic prediction (minimum flow 50.0 and maximum flow 9977.45)

| Model | Without Facebook Data | | | | With Facebook Data | | | |
|---|---|---|---|---|---|---|---|---|
| | RMSE | MAE | MAPE | $R^2$ | RMSE | MAE | MAPE | $R^2$ |
| LSTM | 1328.72 | 962.72 | 93.08 | 0.32 | 1325.69 | 949.17 | 92.43 | 0.33 |
| GCN-LSTM | 1183.41 | 802.02 | 92.50 | 0.45 | 1155.02 | 758.32 | 91.75 | 0.48 |
| CNN-LSTM | 1180.63 | 798.67 | 92.15 | 0.46 | 1134.51 | 756.05 | 88.98 | 0.50 |
| DGCN-LSTM | 1084.36 | 746.18 | 82.94 | 0.55 | 1053.24 | 748.48 | 73.51 | 0.57 |
| DGCN-LSTM (transfer learned) | 514.20 | 328.98 | 23.84 | 0.89 | 393.98 | 276.32 | 13.49 | 0.93 |

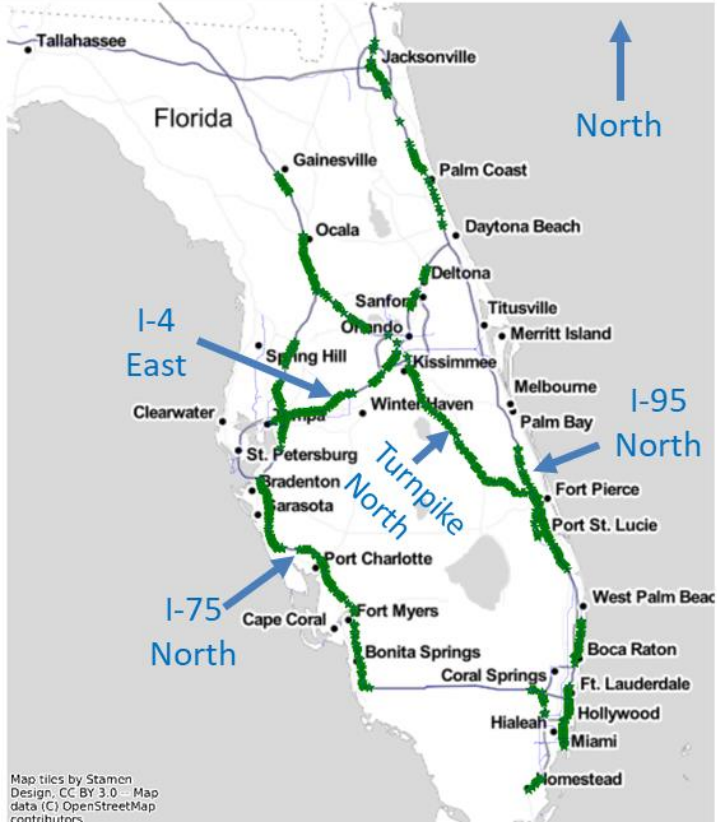

**Fig. 1.** Selected detector distribution from RITIS

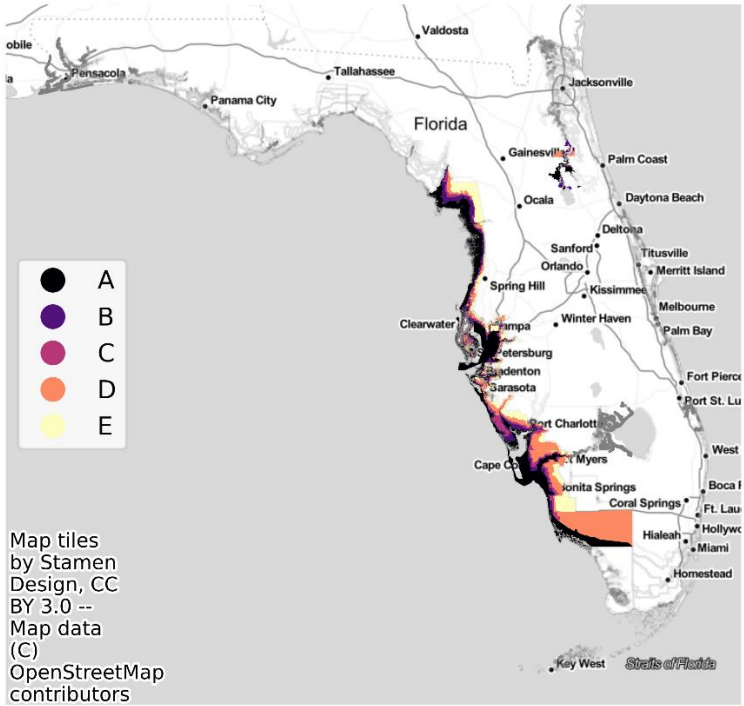

**Fig. 2.** Mandatory evacuation zones during Hurricane Ian

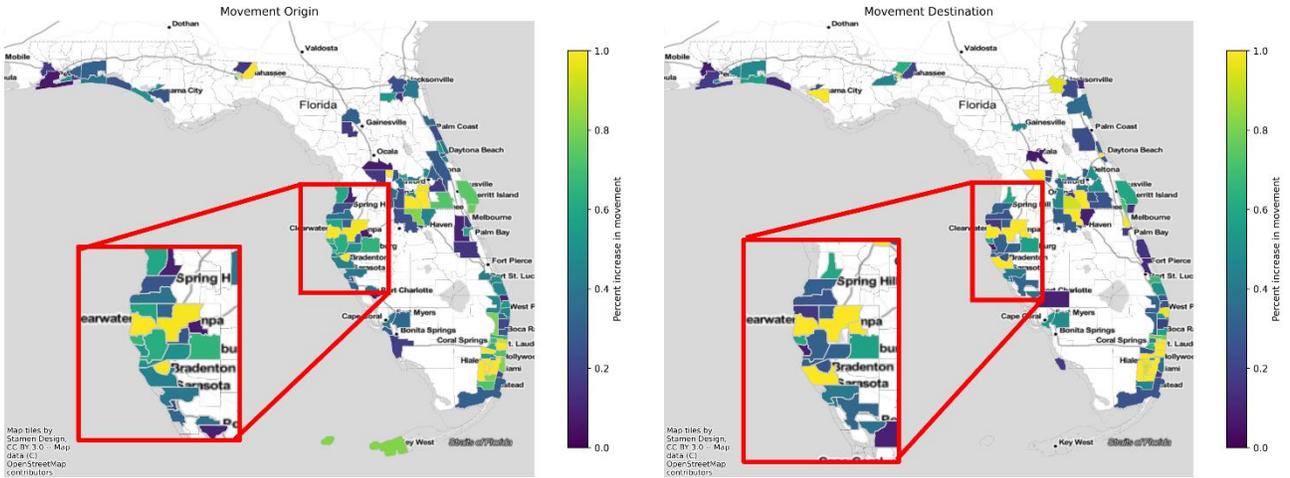

**Fig. 3.** (Left) Origins of movements; (right) destinations of movements during September 26 (3 am-11 am)

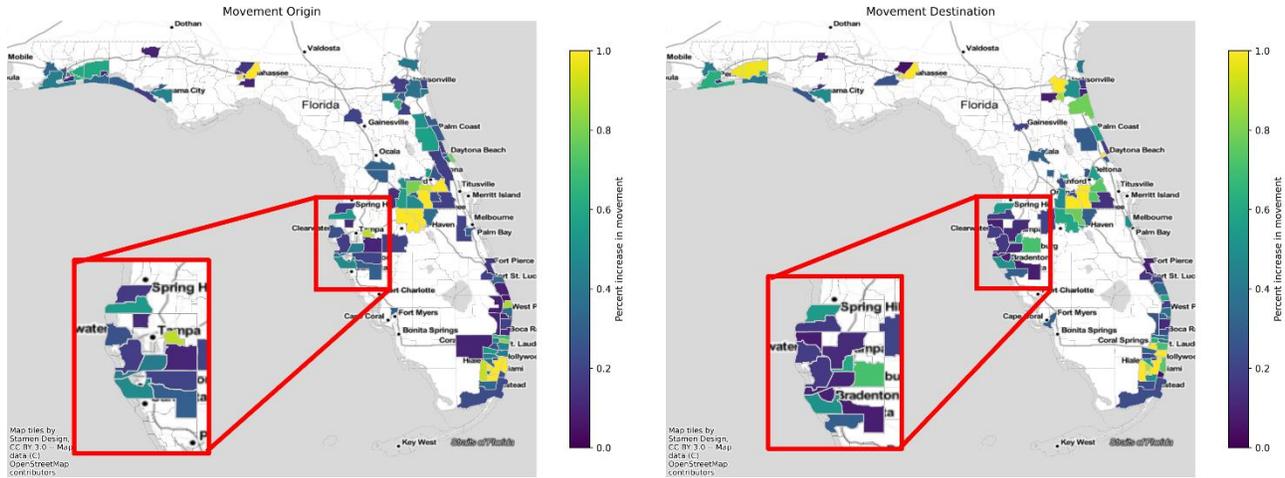

**Fig. 4.** (Left) Origins of movements; (right) destinations of movements during September 27 (3 am-11 am)

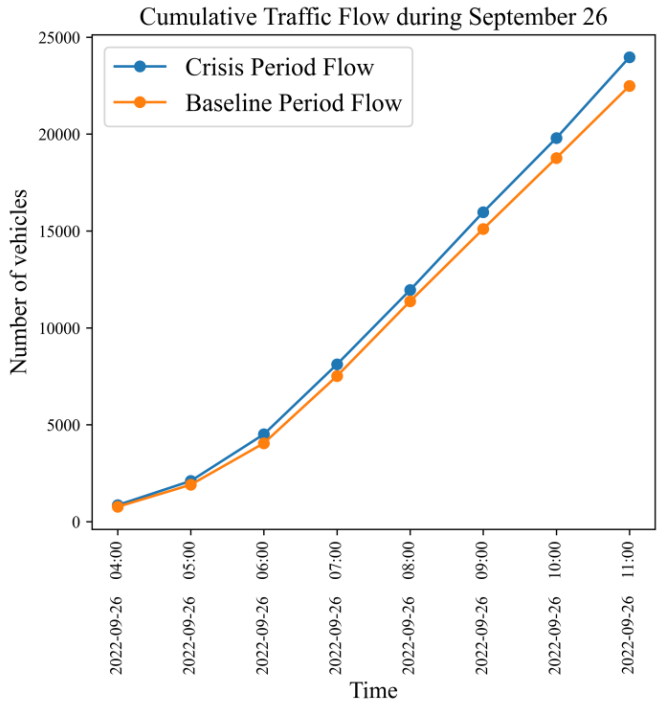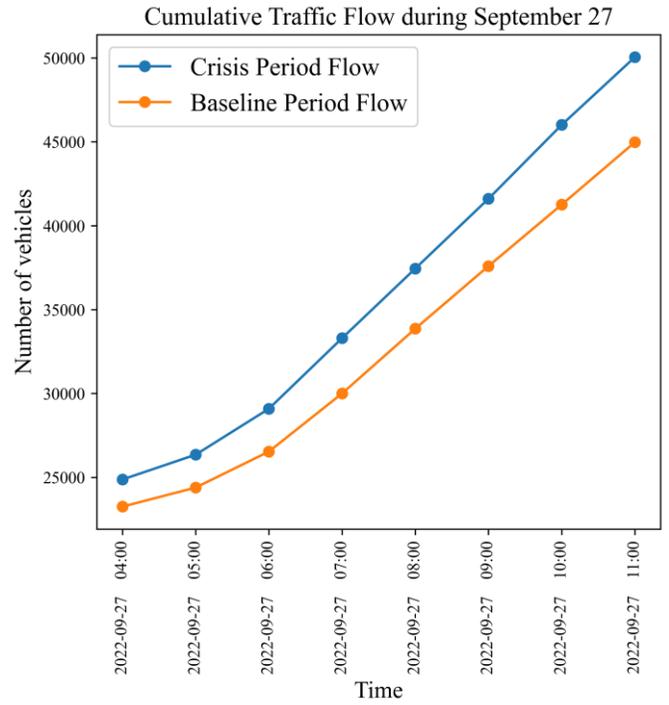

**Fig. 5.** Eastbound cumulative traffic flow on an I-4 detector

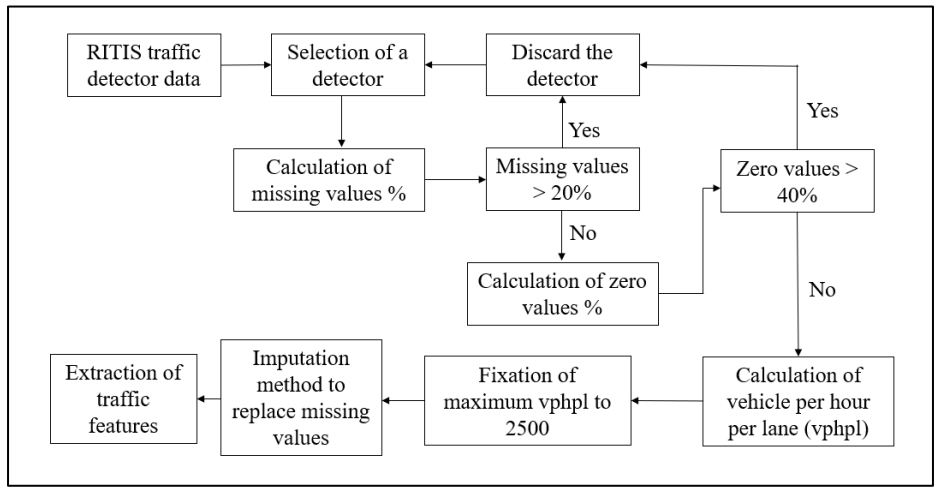

**Fig. 6.** Processing of traffic detector data

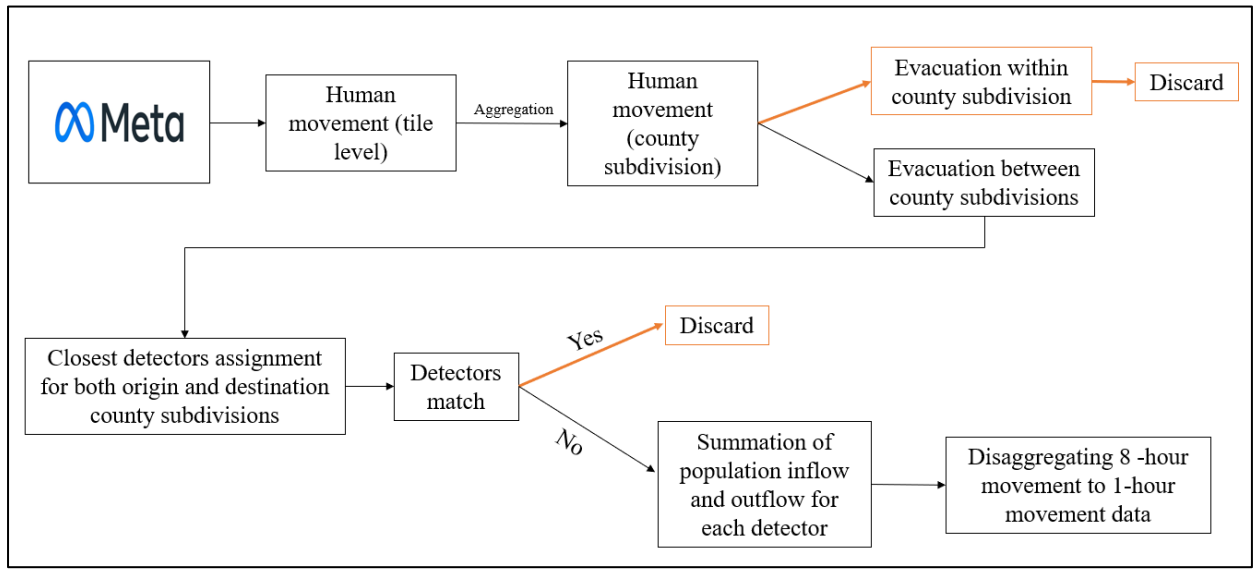

**Fig. 7.** Processing of Facebook movement data

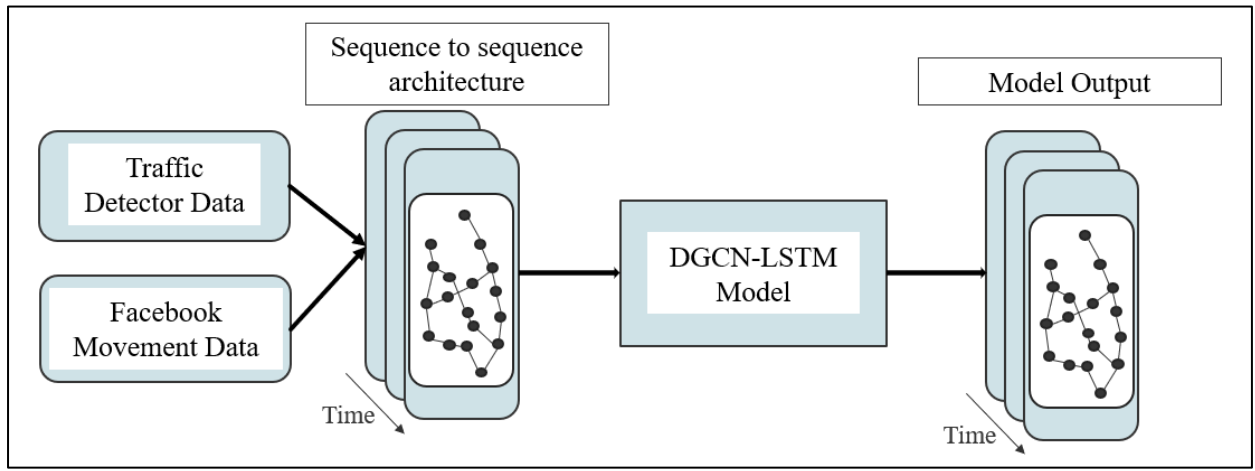

**Fig. 8.** Framework of the DGCN-LSTM model

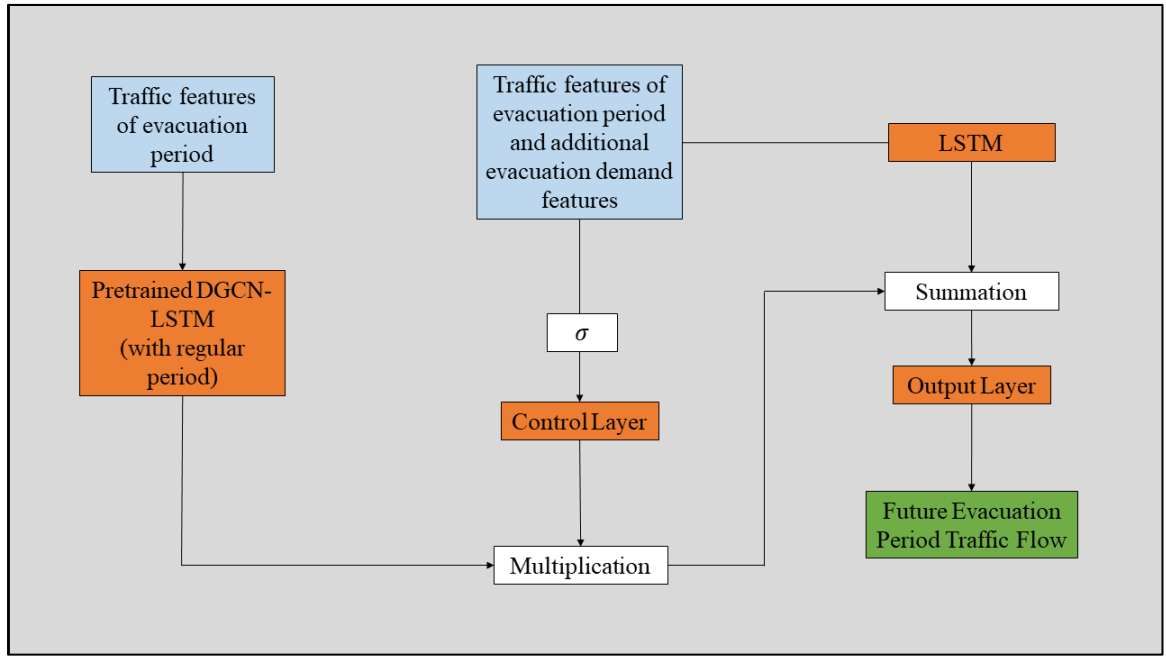

**Fig. 9.** Transfer learning approach for evacuation period traffic prediction

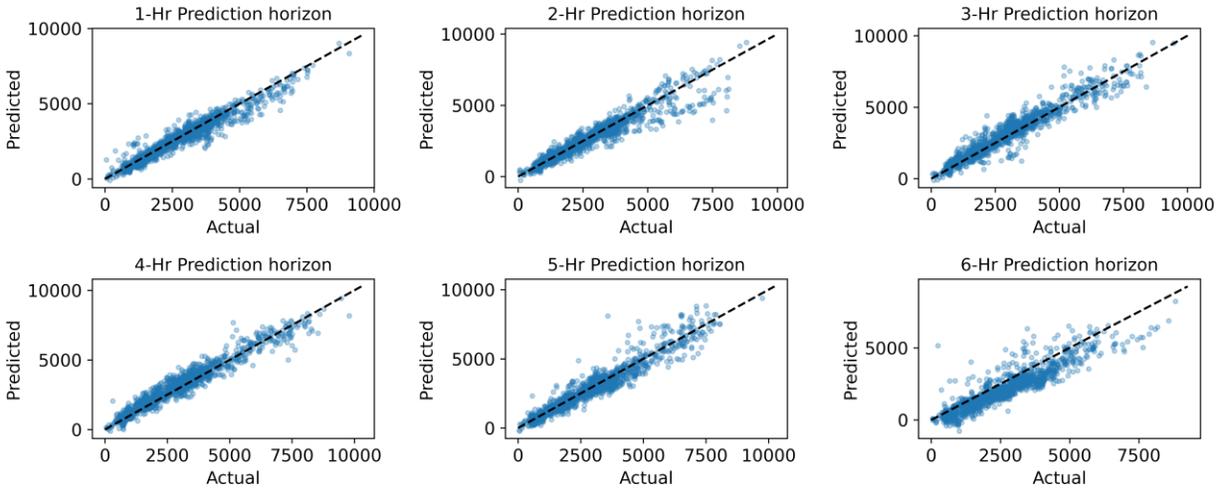

**Fig. 10.** Correlation between actual traffic and predicted traffic for 6-hour time horizon (when Facebook movement data is not used)

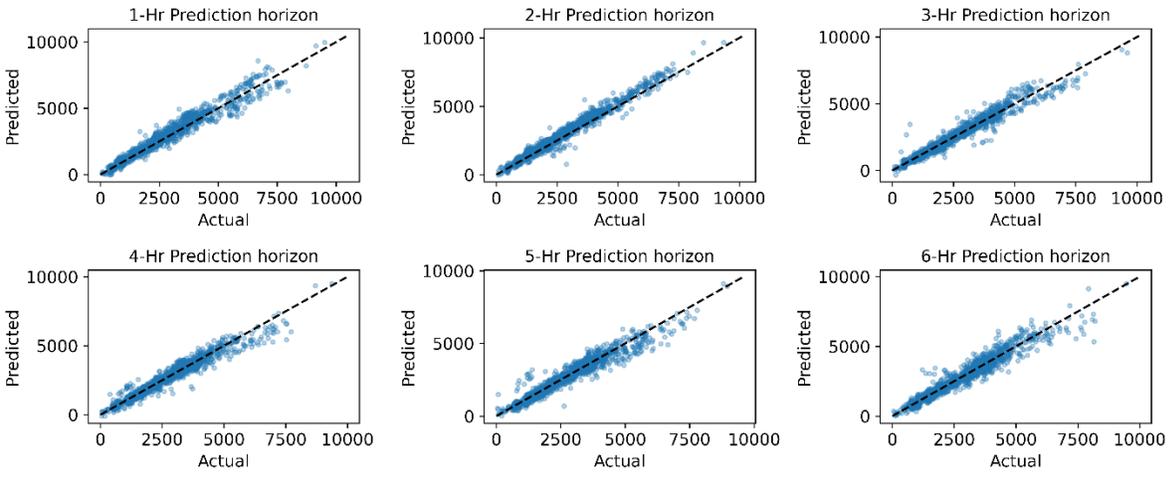

**Fig. 11.** Correlation between actual traffic and predicted traffic for 6-hour time horizon (when Facebook movement data is used)

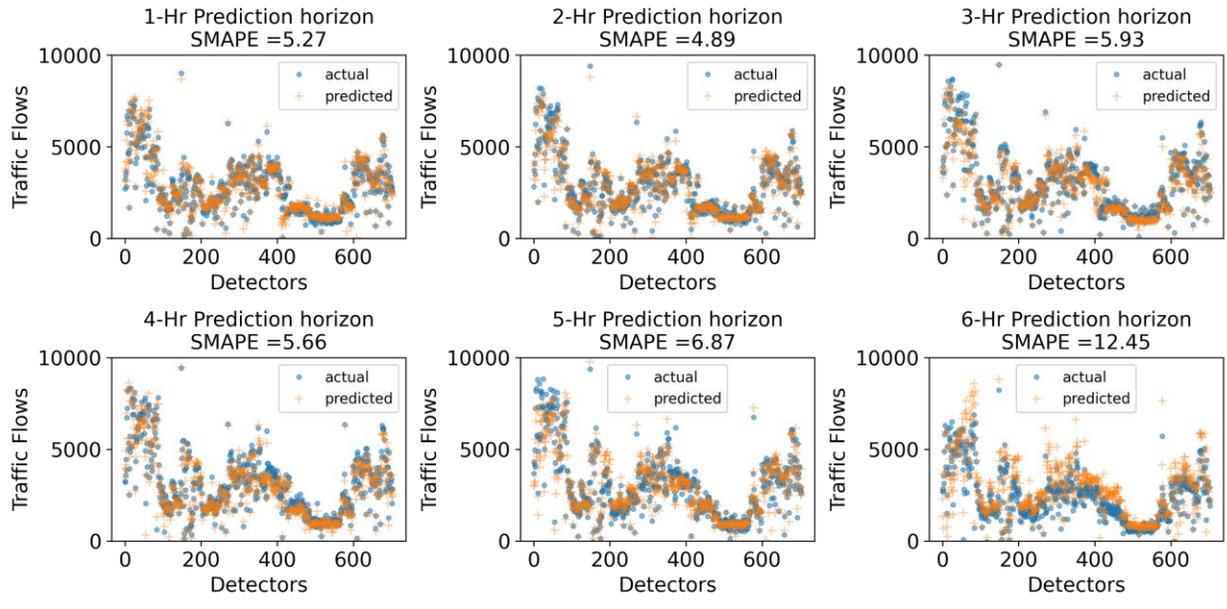

**Fig. 12.** Detector wise actual flow vs. predicted flow with SMAPE values (without Facebook movement data)

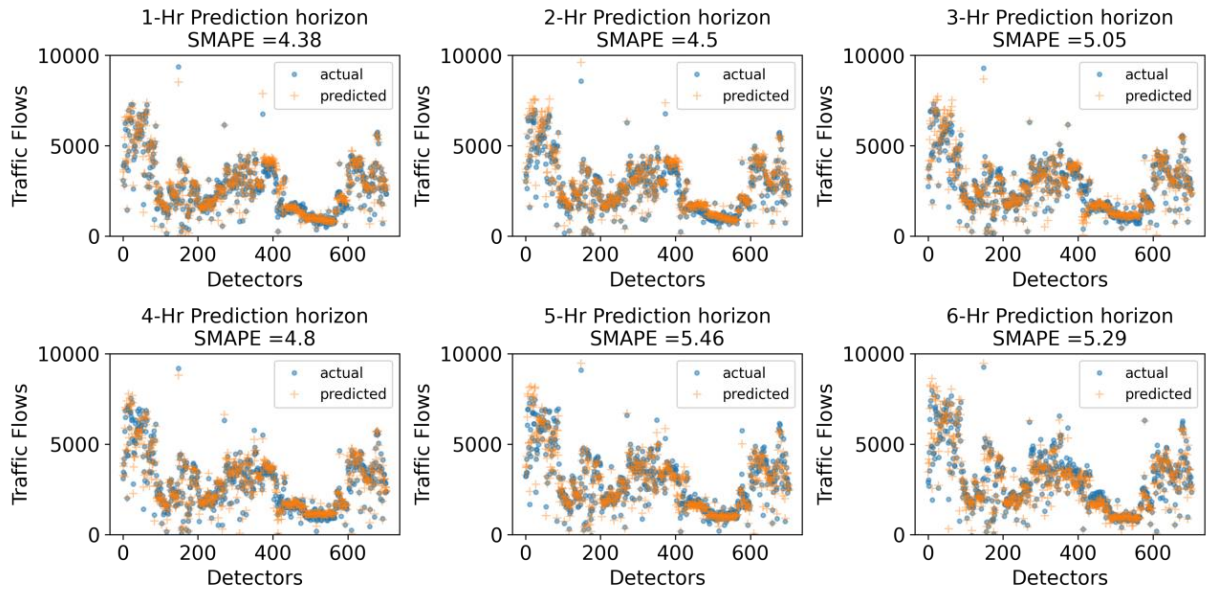

**Fig. 13.** Detector wise actual flow vs. predicted flow with SMAPE values (with Facebook movement data)

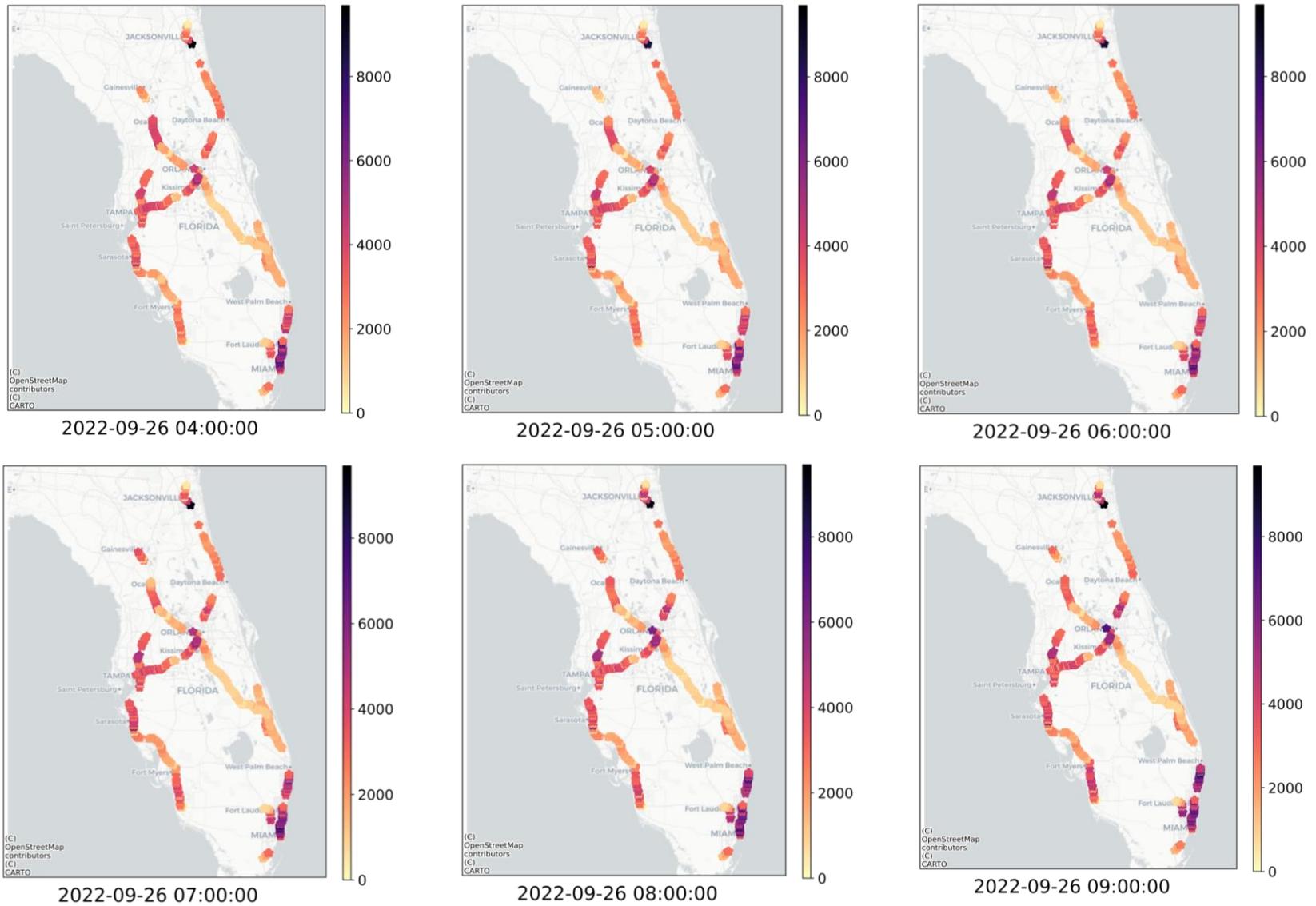

**Fig. 14.** Congestion propagation visualization of actual traffic flow (vertical color bar denotes traffic flow)

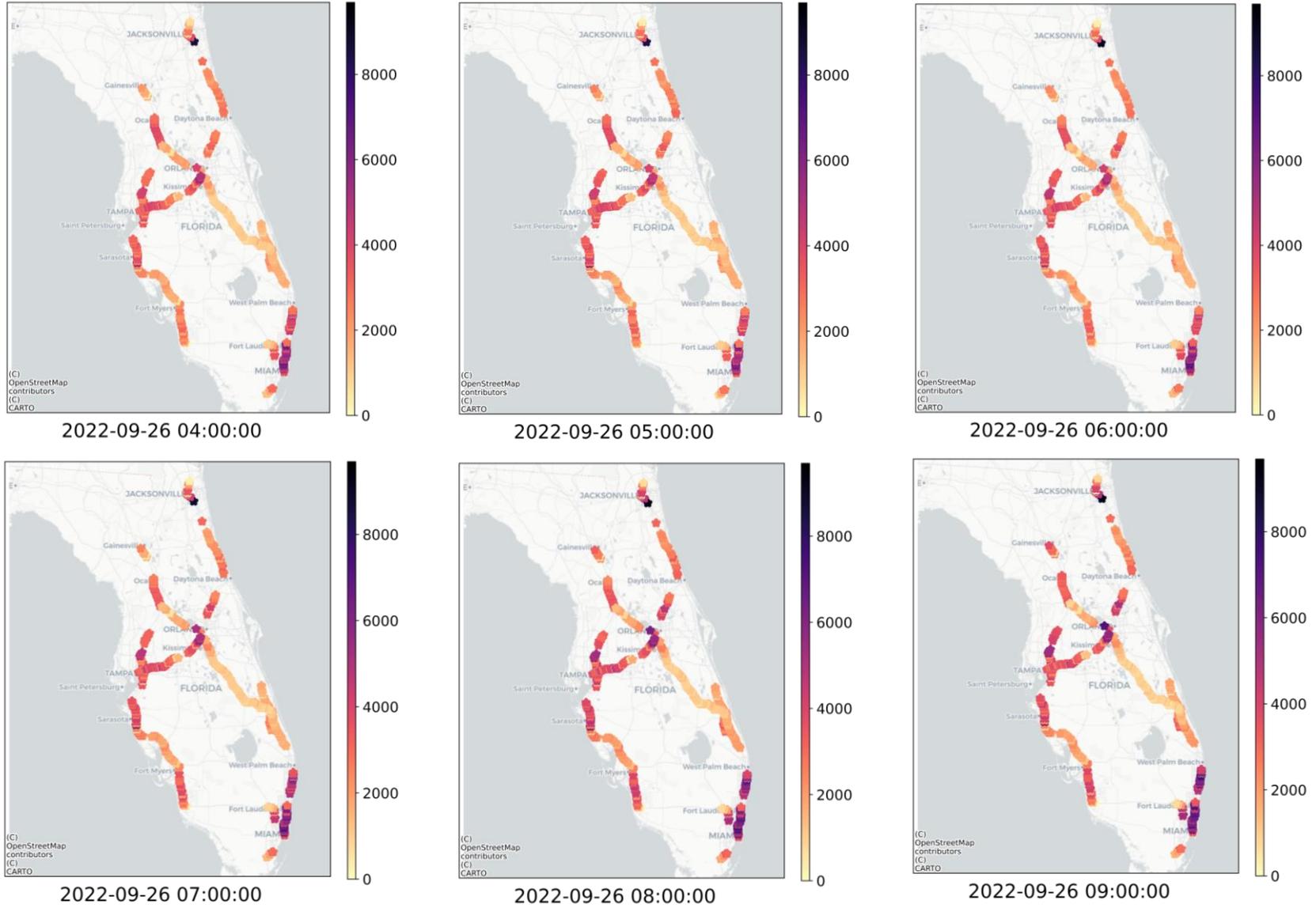

**Fig. 15.** Congestion propagation visualization of predicted traffic flow (vertical color bar denotes traffic flow)